\documentclass[10pt,twocolumn,letterpaper]{article}

\usepackage[pagenumbers]{cvpr}   

\usepackage{graphicx}
\usepackage{amsmath}
\usepackage{amssymb}
\usepackage{booktabs}
\usepackage{blindtext}
\usepackage[pagebackref,breaklinks,colorlinks]{hyperref}
\usepackage[capitalize]{cleveref}
\crefname{section}{Sec.}{Secs.}
\Crefname{section}{Section}{Sections}
\Crefname{table}{Table}{Tables}
\crefname{table}{Tab.}{Tabs.}

\def\cvprPaperID{*****} 
\def\confName{CVPR}
\def\confYear{2023}

\begin{document}
\title{Image Stabilization for Hololens Camera in Remote Collaboration}

\author{Gowtham Senthil $^*$, Siva Vignesh Krishnan $^*$, Annamalai Lakshmanan $^*$,
Florence Kissling
\thanks{G. Senthil and A. Lakshmanan are with the Department of Information Technology \& Electrical Engineering, ETH Z\"urich. S. V. Krishnan is with the Department of Mechanical \& Process Engineering, ETH Z\"urich. F. Kissling is with Department of Computer Science, ETH Z\"urich.\newline
* These authors contributed equally to this work.\newline
E-mail of the corresponding author: sivavigneshk@gmail.com}
}

\maketitle
\begin{abstract}
   With the advent of new technologies, Augmented Reality (AR) has become an effective tool in remote collaboration. Narrow field-of-view (FoV) and motion blur can offer an unpleasant experience with limited cognition for remote viewers of AR headsets. In this article, we propose a two-stage pipeline to tackle this issue and ensure a stable viewing experience with a larger FoV. The solution involves an offline 3D reconstruction of the indoor environment, followed by enhanced rendering using only the live poses of AR device. We experiment with and evaluate the two different 3D reconstruction methods, RGB-D geometric approach and Neural Radiance Fields (NeRF), based on their data requirements, reconstruction quality, rendering, and training times. The generated sequences from these methods had smoother transitions and provided a better perspective of the environment. The geometry-based enhanced FoV method had better renderings as it lacked blurry outputs making it better than the other attempted approaches. Structural Similarity Index (SSIM) and Peak Signal to Noise Ratio (PSNR) metrics were used to quantitatively show that the rendering quality using the geometry-based enhanced FoV method is better. Link to the code repository - \href{https://github.com/MixedRealityETHZ/ImageStabilization}{https://github.com/MixedRealityETHZ/ImageStabilization}

  \textbf{ Keywords - Image Stabilization, Remote Collaboration, Scene Reconstruction, NeRFs}
\end{abstract}
\vspace{-0.7cm}
\section{Introduction}
\label{sec:intro}
\begin{figure}
  \includegraphics[width=\linewidth]{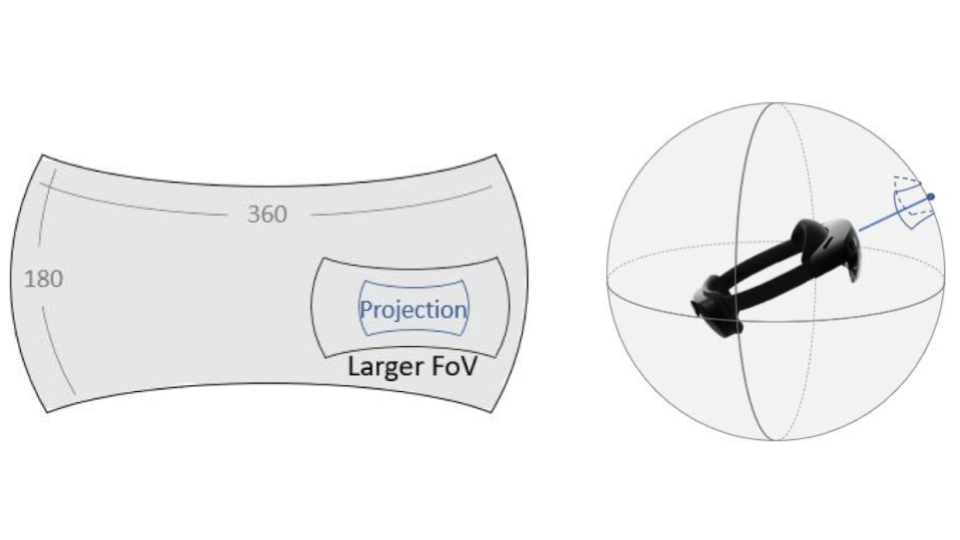}
  \caption{Concept of Enhanced FoV in HoloLens2}
  \vspace{-0.6 cm}
  \label{fig:image_stabilisation}
\end{figure}

AR has become an effective tool for real-time remote assistance and collaboration due to its immersive visual technology. However, the existing AR devices have many drawbacks\cite{problems}, including limited field-of-view ($64.69$\textdegree\ for Microsoft Hololens 2 (HL)\cite{camera}) and motion blur. The narrow FoV negatively impacts the perception of the overall context in a scene. Moreover, during remote collaboration, these devices can provide blurry images due to rapid head movements and this is rather an unpleasant experience for the remote viewer. In this paper, we attempt to address these challenges by providing a more stable experience with a larger FoV as shown in Fig. \ref{fig:image_stabilisation} for the remote viewers of the AR device.

To provide a stable visual experience for remote viewers, we require a realistic and detailed geometric representation of the environment. However, generating a precise and realistic reconstruction of the 3D world in real time is infeasible with the current processing power in the AR headsets. Thus, we resort to a two-stage solution wherein the 3D reconstruction is performed offline, while an enhanced view is rendered from this representation during runtime. We capture a localized RGB-D sequence of the target environment and make use of this dataset to perform offline 3D reconstruction using the following two methods.

\begin{figure*}
  \includegraphics[page=1, width=\textwidth]{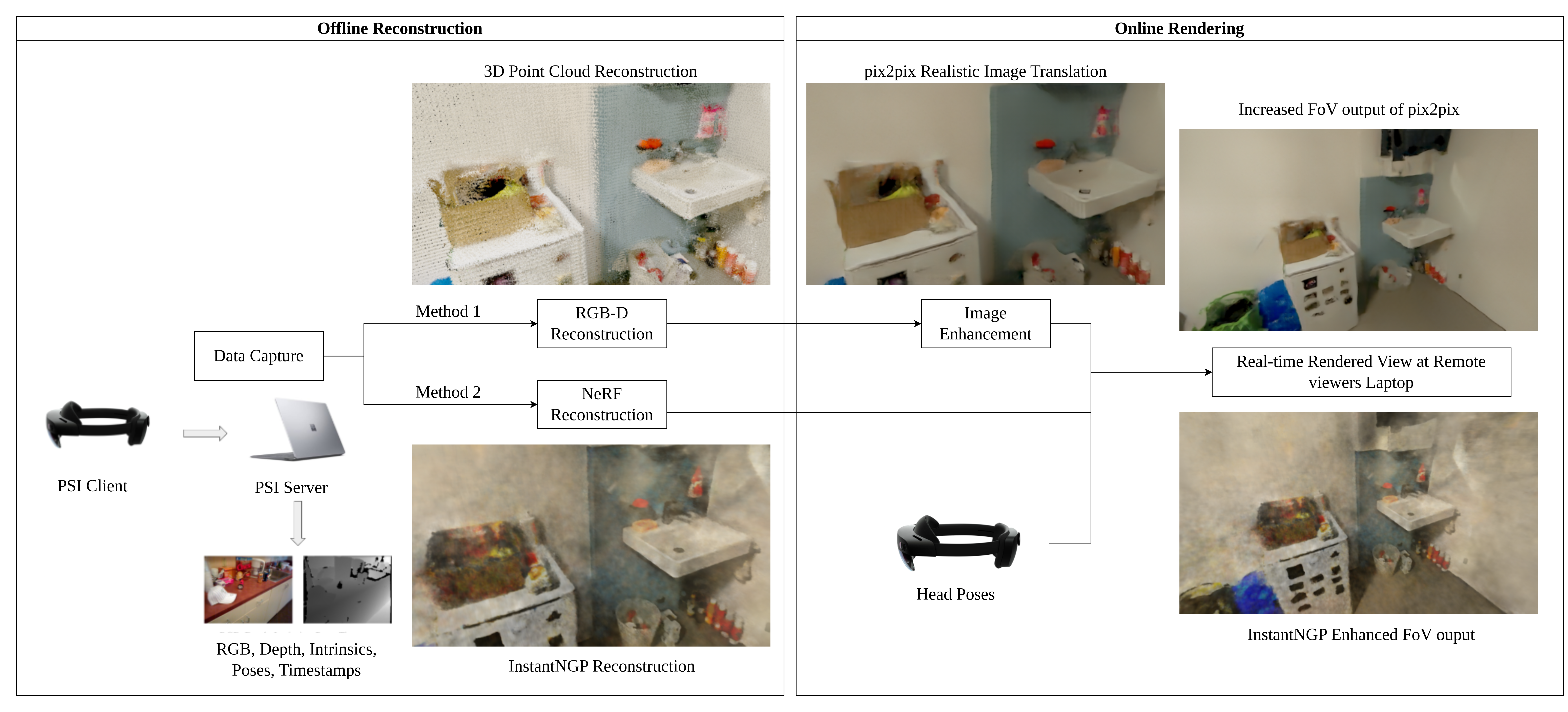}
  \caption{Illustration of the proposed pipeline: Data is captured using HoloLens and PSI framework. In the first method, both RGB and depth images are used to get a 3D point cloud of the scene. Then an image translation model translates the point cloud to realistic-looking images and then an enhanced FoV image is rendered based on the head pose of the user of HoloLens. In the second method, InstantNGP based approach is used to reconstruct a scene and then based on the headposes of the user in realtime, enhanced FoV images are rendered}
    \vspace{-0.6 cm}
  \label{fig:pipeline}
\end{figure*}

In the first method, we use depth information to build a point cloud and mesh representation of the environment. In the second method, we train a deep Neural Radiance Field (NeRF) representation using only RGB images. During runtime, the localized poses of the HL device are transmitted to the remote device, from where we can render the FoV-enhanced images from either of these representations to the remote user. Additionally, we post-process the images obtained from the first approach with image-enhancement techniques to generate realistic-looking images. The proposed pipeline is illustrated in Fig. \ref{fig:pipeline}.

\section{Related Works}
\label{sec:related}
There are several methods for generating a coloured point cloud representation of a scene from multiple RGB images. The most popular Structure-from-motion (SfM) \cite{sfm} involves taking a sequence of images of an object or scene from different viewpoints and using them to reconstruct a 3D model. SfM works by extracting features from the images, such as points or lines, and matching them across different views. The relative positions and orientations of the cameras can then be estimated using these correspondences, and the 3D structure of the scene can be reconstructed. Multi-view Stereo (MVS) \cite{Seitz2006ACA} involves taking multiple images of the same scene from different viewpoints and using them to estimate a dense 3D point cloud with or without calibrated camera poses. MVS works by finding correspondences between image pixels in different views and using them to reconstruct the 3D structure of the scene.

Both SfM and MVS can obtain high-precision estimates in textured areas. However, they face difficulty in feature matching of texture-less areas which are prominent in indoor scenes. This leads to the incompleteness of point clouds or a large number of outliers. Learning-based MVS approaches \cite{https://doi.org/10.48550/arxiv.2109.01129} attempt to overcome this problem with advantages in terms of accuracy and completeness in reconstruction. Methods such as planar priors \cite{Xu_Tao_2020}, multiresolution \cite{Xu2019MultiScaleGC} and depth map completion \cite{9320407} also improve the results of the texture-less region, but they still cannot obtain satisfactory point clouds.

Introducing depth information along with RGB images will simplify the process of 3D reconstruction by a significant amount as discussed in works like 3D reconstruction of a scene with RGB-D information like accurate geometric registration \cite{Choi_2015_CVPR}, joint appearance and geometry optimization \cite{10.1007/s00371-016-1249-5}. 



Since renders of point cloud, despite having a lot of the details, do not look like images captured from the real world. So, we explore techniques to transfer views from the point cloud to the image domain. For this purpose, image-to-image translation and style transfer Generate Adversarial Networks (GANs) can provide good results. There are numerous such networks, both supervised and unsupervised, including some well-known supervised ones \textit{pix2pix} \cite{pix2pix}, \textit{pix2pixHD} \cite{pix2pixHD}, and \textit{pSp} \cite{pSp}. 

While \cite{pix2pix} introduced the concept of generic image-to-image translation, its results were rather blurry and unsuitable for high-resolution images. The follow-up \cite{pix2pixHD} has multiscale generator and discriminator architectures to generate high-resolution images. \cite{spade} introduces a new spatially-adaptive normalization layer to improve the quality of translating semantic maps into high-resolution images. \cite{pSp} encodes the input image into a latent vector which is then passed to a pre-trained Style GAN \cite{stylegan} to obtain the translated image output, and was shown to generate high-quality image outputs. In our work, we demonstrate that the simplest \textit{pix2pix} already provides reasonable quality results for view enhancement.

An alternative for rendering images at very high resolution is Neural Radiance Fields (NeRFs) \cite{Nerf} which recently shot into prominence. The model involves an implicit network of Multi-Layer Perceptron (MLP) that can reconstruct complex 3D scenes. However, training involves several hours if not days while rendering a scene would take around 1 minute. This is highly impractical for our application. Fast-NeRFs \cite{fast-nerf} reduced this inference time to 5ms by efficiently caching the position's map in space. More recent works like NICE-SLAM\cite{nice-slam} and Instant-NGP\cite{instant-ngp} can reconstruct scenes within a few minutes.

While other NeRF approaches work with just RGB images and poses, NICE-SLAM augments the depth information to give an environment with less noise. However, the images don't look real as it uses a hierarchical representation, first reconstructing it sparsely and then doing a fine-level refinement. Instant-NGP uses a multi-resolution hash-based encoding on MLPs and this reduces the reconstruction time to around 5 minutes while the inference time is brought down to 200ms. We proceed to implement the Instant-NGP model because of the faster training and almost real-time inference while also rendering realistic high-quality images.

\section{Proposed Pipeline}
\label{sec:proposed}
\subsection{Data Capture}
\label{subsec:data}
HoloLens2 (HL) has an RGB camera sensor and a range (depth) sensor, essentially, having RGB-D information. We require a sequence of synchronized RGB images and depth maps with known poses for 3D reconstruction. HL's \textit{Research Mode API} \cite{researchmode} exposes this data, which we access through a wired connection using Microsoft's Platform for Situated Intelligence (PSI) \cite{psi} applications. Due to the limited bandwidth of the wired connection, we were limited to recording a reduced resolution of RGB images than the maximum that the hardware supports. We record RGB images at $1280\times720$ RGB images at $25$Hz, $320\times288$ depth maps at $5$Hz, and also their corresponding extrinsic poses and intrinsics.

PSI's HoloLens CaptureApp is deployed on the HL with the above settings, and data is transferred to a wired Windows laptop running the PSI's HoloLens CaptureServer. The user walks around the target environment extremely slowly to ensure images are not very blurry and tries to cover all viewing angles for the entire environment. This takes approximately $5$ minutes for a room of size $10-15\ m^2$. This data is then processed with PSI's HoloLens Exporter to create a \textit{train} dataset which is used to reconstruct the room. The \textit{test} dataset is typically much shorter and contains fast motion and is used for both qualitative and quantitative evaluations.

\begin{figure}
  \includegraphics[width=\linewidth]{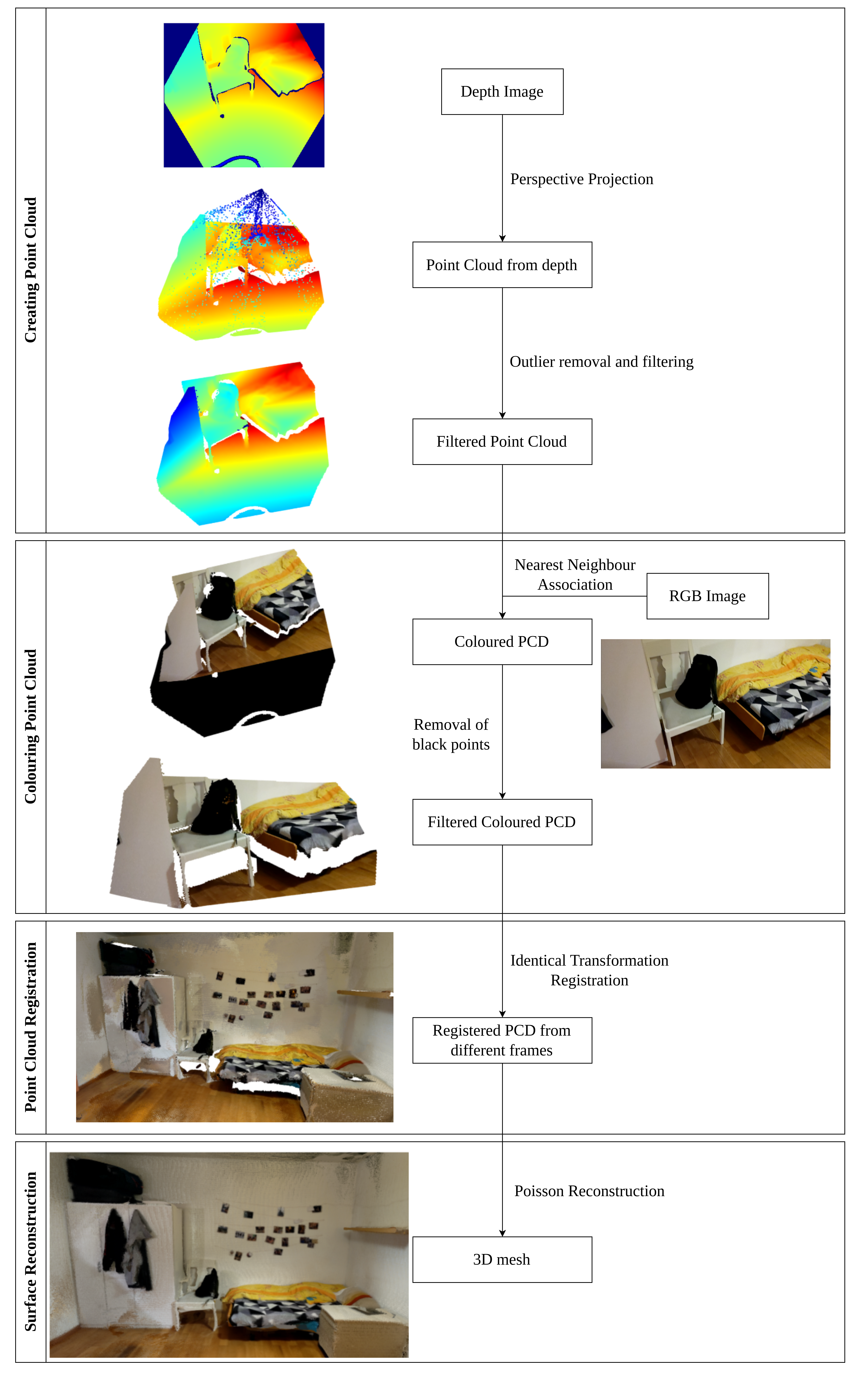}
  \caption{Illustration of the RGB-D mesh reconstruction}
  \label{fig:RGB-D_pipeline}
    \vspace{-0.6 cm}
\end{figure}

\subsection{RGB-D Mesh Reconstruction}
\label{subsec:mesh}
Internally, HoloLens2 is able to output a \textit{SpatialMesh} of the scanned scene. But we do not make use of it as it lacked colour information and as it was not very detailed. Hence, we tried to reconstruct a coloured mesh of the scene for rendering. We use the depth images and RGB images, along with the camera intrinsics, extrinsics (also refered to as poses in this article), and time stamps at which the image is captured to create the point cloud representation of the scene, in our case, a room. We follow the four steps below to do the same. Fig. 
\ref{fig:RGB-D_pipeline} illustrates how a mesh is generated from RGB-D information from HoloLens.

\textit{Step 1: Creating the point cloud.} We create a set of point clouds from each depth image in the dataset. For this, we first undistort the images and for each pixel in the undistorted image, we use the intrinsic parameters such as the focal length and the principal point and the depth value to calculate the 3D position of the point corresponding to that pixel in the camera's coordinate system. We use the Open3D \cite{open3d} library and the location of the points to create a point cloud. Since the point cloud was noisy, we tried to remove them using a radius outlier filter and a statistical outlier filter from Open3D. Using these two filters, we were effectively able to reduce the noise. Depth cameras generally do not work well with highly reflective, transparent, or black surfaces and hence some of the points were very far away from or very close to the camera. This is an inherent problem of the depth sensor as it uses time-of-flight to measure the distance. To overcome this, we used the fact that the data was recorded inside a room and away from various artefacts. Hence, we filtered the points and removed the ones are either very close to or very far away from the camera. Essentially, only the points that lie within a certain depth range were chosen for further processing. The distance range was manually fine-tuned to work well with various frames in various rooms.

\textit{Step 2: Colourizing the point cloud. }For every depth image, we find the associating RGB image and raycast to get the colourized point cloud. Since the depth image and the RGB images are from different sensors, we associate them using the timestamp information employing the nearest neighbour approach. The point cloud obtained from that particular depth image is transformed into a world frame and then projected onto the image plane using the RGB camera intrinsics, and extrinsics, using the standard projective transformation function from OpenCV. We choose the depth points that have some colour assigned to them using the indices. We then colourise the point cloud using Open3D utilities. We remove the points that do not have a colour associated with them from the point cloud. Additionally, we removed the points that are close to the image edges (farther away from the optical center) as the distortion is higher in those regions. By doing so, we reduce the colour noise in the point cloud.

\textit{Step 3: Point Cloud Registration.} Since all the point clouds from depth images are processed in a particular world frame, we stitch them together through identical transformation, after voxel-wise downsampling. We then apply statistical outlier removal once again to remove points that are further away from their neighbours exploiting the fact that the scene is a room. Since the overlapping regions were darker than the non-overlapping regions and since it affected the output of the later part of the pipeline (image translation network training), we stitched the points that were not already present or present very close to existing points, hence avoiding overlap.
The colours of the point clouds were not very detailed for smaller objects present in the room. To address this, we implemented coloured-Interior Closest Point (coloured-ICP) algorithm \cite{park2017colored} for point cloud registration instead of identical transformation to improve the association of colour and reduce the localisation errors of the camera poses. The coloured ICP was removed from the pipeline as we figured it does not improve the scene by even a negligible amount. 

\textit{Step 4: Surface Reconstruction from Point Cloud.} There are many standard algorithms available to create a 3D surface model from a set of 3D points such as Delaunay triangulation \cite{lee1980two}, Ball pivoting \cite{817351}, Poisson Surface Reconstruction \cite{kazhdan2006poisson}, and Isosurface extraction \cite{489388}. Since we require a detailed and accurate 3D model that is similar to the real-world ground truth, we used Poisson surface reconstruction. This method creates a watertight triangular mesh surface model by fitting a polynomial surface to the input points and using an octree data structure to reconstruct the surface. We employ the Open3D utility function to create a 3D mesh. For this, we first estimate the normals for the point cloud by locally fitting a plane per 3D point and then propagating the normal orientation using a minimum spanning tree to ensure consistency. Further, density-based filtering was employed to remove vertices and triangles that have low support from the 3D mesh. The mesh was then filtered to include only the region inside a bounding box defined by the point cloud. In this way, we reduced the triangles that were outside the room.

\begin{figure*}
  \includegraphics[page=1, width=\textwidth]{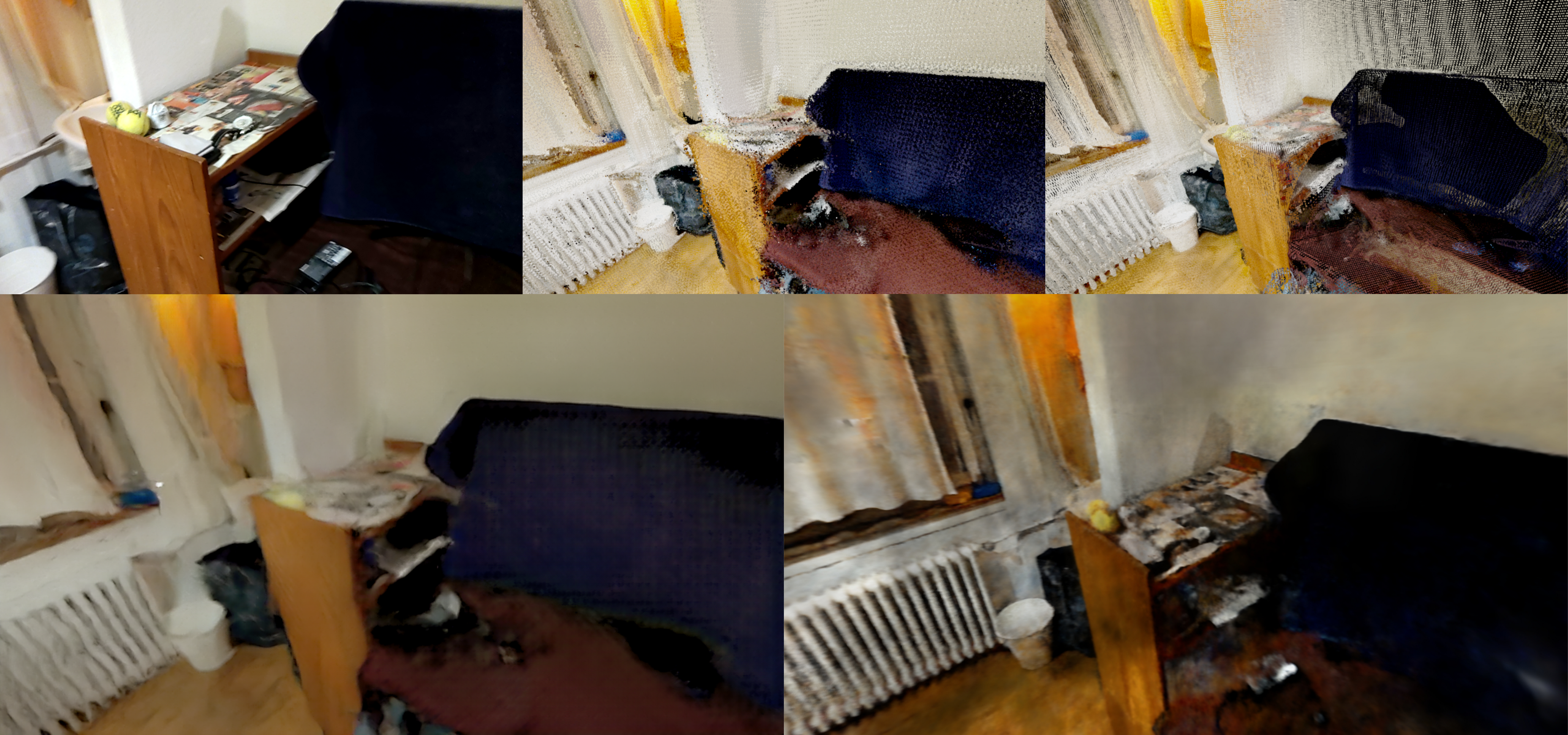}
  \caption{Qualitative comparison of renders with enhanced FoV from different reconstruction techniques for dataset 1.\\ Top Row: Original image, point cloud, Mesh. Bottom Row: point cloud + Pix2pix enhancement, NeRF}
    \vspace{-0.4 cm}
  \label{fig:dataset1}
\end{figure*}

\subsection{NeRF Reconstruction}
\label{subsec:nerf}
For NeRF-based reconstruction, we use the Instant NGP model. Since we do not require depth maps for this model, we consider only the colour images (RGB) and camera poses. 

\textit{Step 1: Selecting high-clarity images.} NeRFs require high-resolution images to give a good representation of a 3D complex scene. As this model is trained with just RGB images, even having a few blurry images can result in a lot of noise in the model. In total 150-200 images are usually required for the reconstruction of a scene. We chose the number of images as 200 and the sharpness threshold as 150. To enforce these constraints, all the images were first checked for their sharpness. Based on their sharpness within a group, the sharpest image is selected. If the sharpness is lesser than the threshold for a certain image, the sharpness is increased to 150 using an unsharp mask. The number of images per group is decided inherently using an image selection function so as to obtain 200 images in total. 

\textit{Step 2: Pose Refinement.} The poses for the selected images are taken and the axis is transformed. Then they are converted to NeRF input-type poses which require a central alignment based on which the images are subject to transformations. Both the train dataset and test dataset are subjected to this pose refinement. The test dataset however does not require a separate central alignment-based transformation as we can use the transformation generated for the train dataset.

\textit{Step 3: Training and Rendering.} Now, the images and the transformed poses of the train dataset are fed to the model which is able to reconstruct a scene. From the scene, we now give the test poses to render the images. Rendering is usually done by ray-casting across all pixels of an image to get the RGB values along the ray which essentially gives out the images required. We also render views with an enhanced FoV of $100^{\circ}$ so as to increase the perspective. Both training and rendering were done using an RTX3090 with 8GB memory as Instant NGP requires a cuda compute capability greater than 7.0 to use Fullyfused MLP (responsible for bringing down the training time). For our scenes, the training time was between $5-7\ min$ and the rendering time per image was observed to be $200\ ms$.

\subsection{View Enhancement}
\label{subsec:enhance}
We use Open3D \cite{open3d} framework to render image views from the point cloud or mesh representations using poses given by HL. However, the rendered images do not look like realistic images (see \cref{fig:dataset1}). We perform the discussed learned view enhancement step to make them look like realistic images. Here, we make use of a standard image-to-image translation architecture, \textit{pix2pix} \cite{pix2pix} which is trained on the \textit{train} dataset. 

During training time, the input to the network is a rendered view of the point cloud or mesh using HL pose, and the corresponding original image is used to train the GAN in a supervised fashion. The input images are cropped to $512\times512$ size and trained with random flips for $100$ epochs with $2\times10^{-4}$ learning rate. The learning rate is linearly reduced step-wise to $0$ over another $100$ epochs. The whole training takes around $36$ hours on an RTX 2080 GPU.

During runtime, the HL poses from \textit{test} dataset are used to render views from point cloud representation with increased FoV of about $100$\textdegree\ (compared to $64.69$\textdegree\ from HL2). This view is passed as input to the network to obtain the realistic image for the current frame. The inference time per frame is $7.4\ ms$ on a RTX 2080 GPU.  

\begin{figure*}
  \includegraphics[page=2, width=\textwidth]{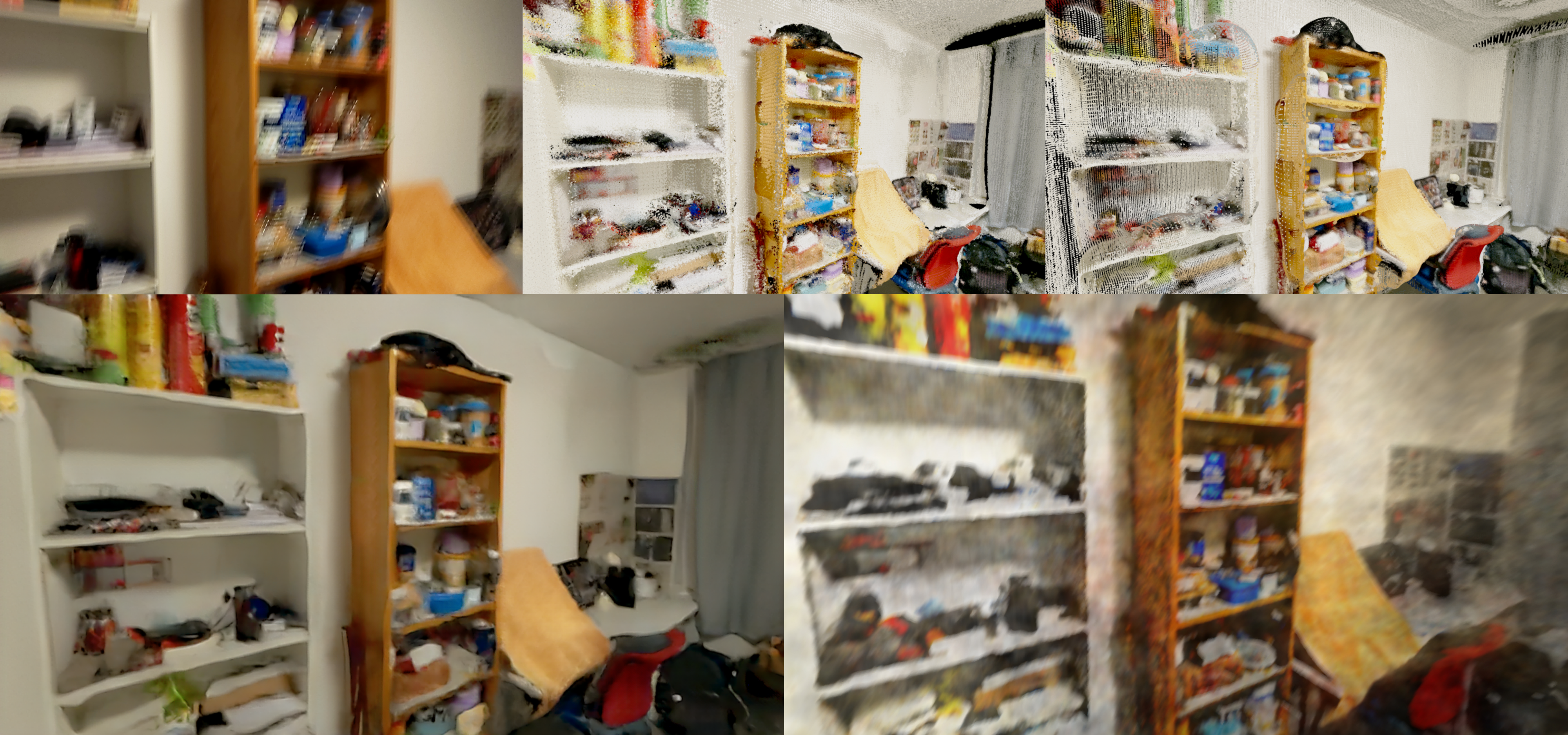}
  \caption{Qualitative comparison of renders with enhanced FoV from different reconstruction techniques for dataset 2.\\ Top Row: Original image, point cloud, Mesh. Bottom Row: point cloud + Pix2pix enhancement, NeRF}
    \vspace{-0.4 cm}
  \label{fig:dataset2}
\end{figure*}

\section{Implementation}
\label{sec:imple}
\subsection{Demo Application}
\label{subsec:api}
We propose an application that deploys the image stabilization pipeline for the purposes of remote assistance. The application consists of a component on the HL and one on a Windows laptop. 

On the HL side, the local user enables Research Mode such that the app can obtain the current head pose of the device. The app streams the live sequence of head poses over a TCP connection to the remote viewer's Windows laptop.

On the Windows laptop, a script listens to this TCP stream. Whenever a packet with a new pose arrives, the script will feed the pose to either the trained Instant-NGP model or the point cloud representation to obtain an enhanced image of the current HL view on the remote laptop. Thus the remote viewer can experience a better viewing quality in real-time given the 3D reconstruction that was performed offline.

Due to time constraints, we did not implement the proposed application. Nevertheless, we believe a real-time solution is feasible as the network delays introduced by our proposed application would be significantly lower than streaming the entire images from HL to the remote viewer.

\subsection{Experimental Results}
\label{subsec:results}
\begin{table}[]
\scalebox{0.85}{
\begin{tabular}{@{}c|cccc@{}}
\toprule
Dataset & point cloud & Mesh & \begin{tabular}[c]{@{}c@{}}point cloud\\ + pix2pix\end{tabular} & NeRF \\ \midrule
Room1   & $11.71\ |\ 0.31$ & $10.83\ |\ 0.15$ & $\mathbf{16.39}\ |\ \mathbf{0.67}$ & $12.18\ |\ 0.56$ \\
Room2   & $11.73\ |\ 0.30$ & $10.77\ |\ 0.16$ & $\mathbf{17.24}\ |\ \mathbf{0.70}$ & $14.30\ |\ 0.65$ \\ 
\bottomrule
\end{tabular}}
\caption{Quantitative comparison of renders of 2 test environments from different reconstruction techniques (PSNR (dB) $|$ SSIM)}
\vspace{-0.6 cm}
\label{tab:metrics}
\end{table}
We capture two environments, each of which is an approximately $10-12\ m^2$ room, and capture a \textit{train}, and \textit{test} sequence for each room. We use the \textit{train} dataset to train the NeRF network, perform RGB-D point cloud reconstruction, and train the view enhancement network. The train dataset Room1 after 3D reconstruction has 12418960 points in the point cloud and 2600068 vertices in the 3D mesh. The train dataset for Room2 after 3D reconstruction has 14236652 in the point cloud set and 2561906 vertices in the mesh. It is to be noted that the reconstruction of 3D coloured point cloud for a frame 

We report the PSNR and SSIM \cite{ssim} metrics of each representation's rendering (with the same FoV as HL) against the original images of \textit{test} dataset in \cref{tab:metrics}. While these are good metrics for image comparison and help us in evaluating the results, it is to be noted that the original ground-truth images from HL have a lot of motion blur for this dataset as it consists rapid motion. Hence, the numbers are not completely representative of the image quality.

The point cloud and mesh representations have limited details and do not resemble real images and hence have poor PSNR and SSIM scores on both datasets. The pix2pix enhancement improves PSNR by upto $5.5dB$ and SSIM more than doubles and renders more accurate images. The NeRF-based approach scores lower than pix2pix-enhanced renders, despite having more details on the background and a sharper image. This is mainly due to the smokey nature (as seen in \cref{fig:dataset2} of the NeRF reconstruction which in turn is due to the blurry images used to train the model.

In \cref{fig:dataset1}, the limited FoV of HL leads to missing out on the overall context of the room. This is solved by rendering views with more than $1.5$ times the original FoV. In \cref{fig:dataset2}, we can see a simple case of motion blur from HL, which is even worse in general. Since our renderings are done frame-by-frame from offline reconstructions, we don't face the issue of motion blur and retain most of the detail of the scene. In both figures, it is clear that the point cloud or mesh representation does not resemble real-world lighting, despite having most of the details. NeRF representation while accurately representing the background is too noisy to be usable.

\section{Limitations and Future Scope}
\label{sec:limits}
Data transfer rates from HL to a wired Windows laptop were limited, and we could only transfer RGB images at a lower resolution and framerate than the maximum supported. This also meant lower resolution for training the DNNs, affecting the reconstruction quality and the viewer experience. In addition, we found that individual frames from the HL's video stream had lot of motion blur despite very slow movement. This greatly reduced the quality of NeRF reconstruction and supervised image-to-image translation and resulted in either smokey or blurry outputs. An alternative to solve these issues is to capture high-resolution localized images (instead of video) from the HL and use them for reconstruction, which should provide better-quality output. 

The quality of the point cloud reconstruction appears to be limited by the accuracy of the depth camera in the HoloLens as the data is noisy and they fail catastrophically near edges and on objects or surfaces that are highly reflective or black. Additionally, the depth cameras cannot handle moving objects. For the former, we improved the point cloud by removing the outliers and removing the points that are very close and very far away from the camera. It is to be noted that those distance parameters were hand-tuned exploiting the fact that the scene reconstructed is a room with static elements. The same problem could be tackled more intelligently, in the sense that one could use the RGB image corresponding to the depth image and use deep learning approaches to either get a depth prior \cite{monocu} or fit planes \cite{Mertan_2022} to get more smooth depth values on highly reflective or black plane surfaces. We have ignored the latter as it is out of the scope of this project where we are working with static scenes.

Another factor that limits the quality of the scene reconstruction is the detailing in colour. To create a realistic and accurate 3D reconstruction, it is important to have high-quality colour information for the points in the point cloud. This can be difficult to obtain, especially if the points are far away from the camera (loss of detail) or if the lighting conditions are poor or changing. Invariant illumination could be tackled by faster sensor motion \cite{rs14153551}. There are other factors that diminish the quality of colours in the reconstructed scenes such as: 1. the geometric 3D model being noisy and inaccurate; 2. the RGB camera not being in perfect correspondence with the depth camera, thereby increasing the misalignment of the projected images; 3. the RGB images suffering from blurring and ghosting. To address these sources of errors, one could try to maximize the photometric consistency by jointly optimizing for the camera poses and non-rigid correction function for all images as in \cite{colour_map}.

Processing the point cloud for a room is another bottleneck due to its size. In order to increase the processing speed, one could employ feature-based down sampling approaches \cite{rs12071224}. For example, downsample less in the region where there are many features, and downsample more in the textureless, planar regions (such as a wall). 

The view enhancement \textit{pix2pix} network generates high-quality results for the cropped image size it was trained on. However, when testing on full-resolution image sizes, it results in blurry outputs with less detailing. This can be resolved by either training on full-resolution images which in turn would require more GPU memory and training time or trying out more recent supervised image-to-image translation architectures such as \textit{pix2pixHD} \cite{pix2pixHD}, \textit{SPADE} \cite{spade}, and \textit{pSp} \cite{pSp}. These tackle the problem of fewer details by working at multiple scales to obtain high-resolution images.

The major disadvantage of a NeRF model is the presence of noise clouds in any scene. While the NeRF model \cite{instant-ngp} is able to generate seemingly high-quality images, the amount of noise here is higher. This can be mainly attributed to the 8-megapixel camera of Hololens 2, the low-resolution captured data and the blurriness of the images. One way to reduce the noise while not compromising the image quality was discussed in NeRF-SLAM \cite{nerf-slam} which uses depth information in the Instant-NGP model to reduce the noise.

\section{Conclusion}
\label{sec:conc}
To address the problem of image stabilisation for remote collaboration using AR devices, we proposed a two-stage pipeline. The solution includes an offline 3D reconstruction of an indoor environment, followed by enhanced and smoothed renderings of the reconstructed scene during real-time collaboration. Offline reconstruction is chosen because the current AR headsets do not provide high-computational requirements. For the 3D reconstruction, a geometry-based and a fast NeRF-based (InstantNGP) methods were employed. Experimental results highlight that the rendered views using these approaches can provide a smooth experience for the remote viewer while offering more contextual awareness of the environment than the actual AR device output. The rendering times for both approaches are small and can be done in real-time during remote collaboration. Geometry-based reconstruction with image translation for rendering was found to be better than the NeRF-based method because of faster rendering and lack of noise.

\section{Acknowledgements}
\label{sec:ack}
We thank Taein Kwon, Dr. Mahdi Rad, Dr. Iro Armeni, Prof. Marc Pollefeys for their support and guidance throughout the course of this project . This work was carried out for the course "Mixed Reality" offered during Autumn 2022 semester at ETH Z\"urich and in collaboration with Microsoft.

{\small

\bibliographystyle{unsrt}
\bibliographystyle{ieee_fullname}
\bibliography{egbib}
}
\end{document}